\documentclass[11pt,a4paper]{article}
\usepackage[hyperref]{acl2018}
\usepackage{times}
\usepackage{latexsym}
\usepackage{adjustbox}
\usepackage{multirow}
\usepackage{url}

 \aclfinalcopy % Uncomment this line for the final submission
\title{Comparative Analysis of Neural QA models on SQuAD}

\author{
Soumya Wadhwa  \qquad  Khyathi Raghavi Chandu  \qquad  Eric Nyberg\\
\\
Language Technologies Institute, Carnegie Mellon University \\
\tt{\{soumyaw, kchandu, en09\}@andrew.cmu.edu}
}

\date{}

\begin{document}
\maketitle
\begin{abstract}
The task of Question Answering has gained prominence in the past few decades for testing the ability of machines to understand natural language. Large datasets for Machine Reading have led to the development of neural models that cater to deeper language understanding compared to information retrieval tasks. Different components in these neural architectures are intended to tackle different challenges. As a first step towards achieving generalization across multiple domains, we attempt to understand and compare the peculiarities of existing end-to-end neural models on the Stanford Question Answering Dataset (SQuAD) by performing quantitative as well as qualitative analysis of the results attained by each of them. We observed that prediction errors reflect certain model-specific biases, which we further discuss in this paper.
\end{abstract}

\section{Introduction}
Machine Reading is a task in which a model reads a piece of text and attempts to formally represent it or performs a downstream task like Question Answering (QA). Neural approaches to the latter have gained a lot of prominence especially owing to the recent spur in developing and publicly releasing large datasets on Machine Reading and Comprehension (MRC). These datasets are created from different underlying sources such as web resources in MS MARCO \cite{nguyen2016ms}; trivia and web in QUASAR-S and QUASAR-T \cite{dhingra2017quasar}, SearchQA \cite{dunn2017searchqa}, TriviaQA \cite{joshi2017triviaqa}; news articles in CNN/Daily Mail \cite{chenthorough}, NewsQA \cite{trischler2016newsqa} and stories in NarrativeQA \cite{kovcisky2017narrativeqa}. Another common source is large unstructured text documents from Wikipedia such as in SQuAD \cite{rajpurkar-EtAl:2016:EMNLP2016}, WikiReading \cite{hewlett-EtAl:2016:P16-1} and WikiHop \cite{welbl2017constructing}. These different sources implicitly affect the nature and properties of questions and answers in these datasets. Based on the dataset, certain neural models capitalize on these biases while others are unable to. The ability to generalize across different sources and domains is a desirable characteristic for any machine reading system. Evaluating and analyzing systems on QA tasks can lead to insights for advancements in machine reading and natural language understanding, and \citet{penas2011overview} have also previously worked on this.

One of the first large MRC datasets (over 100k QA pairs) is the Stanford Question Answering Dataset (SQuAD) \cite{rajpurkar-EtAl:2016:EMNLP2016}. For its collection, different sets of crowd-workers formulated questions and answers using passages obtained from $\sim$500 Wikipedia articles. The answer to each question is a span in the given passage, and many effective neural QA models have been developed for this dataset. Our main focus in this work is to perform comparative subjective and empirical analysis of errors in answer predictions by four top performing models on the SQuAD leaderboard\footnote{\url{https://rajpurkar.github.io/SQuAD-explorer/}}. 

We focused on Bi-Directional Attention Flow (BiDAF) \cite{seo2016bidirectional}, Gated Self-Matching Networks (R-Net) \cite{wang2017gated}, Document Reader (DrQA) \cite{chen2017reading}, Multi-Paragraph Reading Comprehension (DocQA) \cite{clark2017simple}, and the Logistic Regression baseline model \cite{rajpurkar-EtAl:2016:EMNLP2016} %which are in positions XXXX respectively on the leaderboard. 
We mainly choose these models since they have comparable high performance on the evaluation metrics and it is easy to replicate their results due to availability of open source implementations. While we limit ourselves to in-domain analysis of the performance of these models on SQuAD in this paper, similar principles can be used to extend this work to study biases of combinations of different models on different datasets and thereby understand the generalization capabilities of these neural architectures.
%Our grand vision is to perform a meta evaluation of the neural QA models on different datasets.

The organization of the paper is as follows. Section \ref{sec:related} gives a comprehensive overview of the models that are compared in further sections. Section \ref{sec:exp} describes the different experiments we conducted, and discusses our observations. In Section \ref{sec:concl}, we summarize our main conclusions from this work and describe our vision for the future.

\section{Relevant Neural Models}
\label{sec:related}

We present a brief overview of the models which we considered for our analysis in this section.

\paragraph{Bi-Directional Attention Flow (BiDAF):} 
This model, proposed by \citet{seo2016bidirectional}, is a hierarchical multi-stage end-to-end neural network which takes inputs of different granularity (character, word and phrase) to obtain a query-aware context representation using memory-less context-to-query (C2Q) and query-to-context (Q2C) attention. This representation can then be used for different final tasks. Many versions of this model (with different types of input features) exist on the SQuAD leaderboard, but the basic architecture\footnote{\tiny{\url{https://allenai.github.io/bi-att-flow/}}} (which we use for our experiments in this paper) contains character, word and phrase embedding layers, followed by an attention flow layer, a modeling layer and an output layer.

\paragraph{Gated Self-Matching Networks (R-Net):}
This model, proposed by \citet{wang2017gated}, is a multi-layer end-to-end neural network whose novelty lies in the use of a gated attention mechanism so as to give different levels of importance to different passage parts. It also uses self-matching attention for the context to aggregate evidence from the entire passage to refine the query-aware context representation obtained. The architecture contains character and word embedding layers, followed by question-passage encoding and matching layers, a passage self-matching layer and an output layer. The implementation we used\footnote{\tiny{\url{https://github.com/HKUST-KnowComp/R-Net}}} had some minor changes for increased efficiency.

\paragraph{Document Reader (DrQA):}
This model, proposed by \citet{chen2017reading}, focuses on answering open-domain factoid questions using Wikipedia, but also performs well on SQuAD (skipping the document retrieval stage). Its implementation\footnote{\tiny{\url{https://github.com/facebookresearch/DrQA}}} has paragraph and question encoding layers, and an output layer. The paragraph encoding is computed by representing each context as a sequence of feature vectors derived from tokens: word embedding, exact match with question word, POS/NER/TF and aligned question embedding, and passing these as inputs to a recurrent neural network. The question encoding is obtained by using word embeddings as inputs to a recurrent neural network.

\paragraph{Multi-Paragraph Reading Comprehension (DocQA):}
This model, proposed by \citet{clark2017simple}, aims to answer questions based on entire documents (multiple paras) rather than specific paragraphs, but also gives good results for SQuAD (considering the given paragraph as the document). The implementation\footnote{\tiny{\url{https://github.com/allenai/document-qa}}} contains input, embedding (character and word-level), pre-processing (shared bidirectional GRU between question and passage), attention (similar to BiDAF), self-attention (residual) and output (bidirectional GRU and linear scoring) layers.

\paragraph{Logistic Regression (LR):}
This model was proposed as a baseline in the SQuAD dataset paper \cite{rajpurkar-EtAl:2016:EMNLP2016} and uses features based on n-gram frequencies, lengths, part-of-speech tags, constituency and dependency parse trees of questions and passages as inputs to a logistic regression classifier\footnote{\tiny{\url{https://worksheets.codalab.org/worksheets/0xd53d03a48ef64b329c16b9baf0f99b0c/}}} to predict whether each constituent span is an answer or not.

\section{Experiments and Discussion}
\label{sec:exp}

We trained the aforementioned end-to-end neural models and compare their performance on the SQuAD development set which contains 10,570 question-answer pairs based on Wikipedia articles.

\subsection{Quantitative Analysis}
To perform a systematic comparison of errors across different models, we investigate the predictions based on the following criteria.

\subsubsection{Span-Level Performance}
The span-level performance is measured typically by Exact Match (EM) and F1 metrics which are reported with respect to the ground truth answer spans. These results are summarized in Table \ref{eval-expt}. The DocQA model gives the best overall performance which aligns well with our expectation, owing to the usage of and improvements in the prior mechanisms introduced in BiDAF and R-Net.

\begin{table}[ht!]
\centering
\resizebox{\linewidth}{!}{
\begin{tabular}{|l|lllll|}
\hline
\bf{Model} & \bf{BiDAF} & \bf{R-Net} & \bf{DrQA} & \bf{DocQA} & \bf{LR}  \\ \hline
\bf{EM (\%)} &  67.67  &  70.12  &  66.00  &  \textbf{71.60} & 40.14 \\
\bf{F1 (\%)}  &  77.31  &  78.94  &  76.28  &  \textbf{80.78} & 50.98 \\
\bf{Correct Sentence (\%)} &  91.05  &  92.37  &  92.40  &  \textbf{93.77} & 83.30 \\
\hline
\end{tabular}
}
\caption{Span and Sentence Level Performance}
\label{eval-expt}
\end{table}

\subsubsection{Sentence-Level Performance}
To investigate trends at different granularities, we also measure sentence retrieval performance. The context given for each question-answer pair is split into sentences using the NLTK sentence tokenizer\footnote{\tiny{\url{http://www.nltk.org/api/nltk.tokenize.html}}}, and the sentence-level accuracy of each of the models is computed (Table \ref{eval-expt}). Since the default sentence tokenizer for English in NLTK is pre-trained on Penn Treebank data which contains formal language (news articles), we expect it to perform reasonably well on Wikipedia articles too. We observe that all the models have high sentence-level accuracy, with DocQA outperforming the other models with respect to this metric as well. Interestingly, DrQA performs better on sentence retrieval accuracy than both BiDAF and R-Net, but has a worse span-level exact match score, which is probably because of the rich feature vector representation of the passage due to the model's focus on open domain QA (and hence retrieval). But, none of these neural models have near-perfect ability to identify the correct sentence, and  $\sim$90\% accuracy indicates that even if we have a perfect answer selection method, this is the best EM score we can achieve. However, incorrect span identification contributes more to errors in prediction for all the models, as seen from the disparity between the sentence-level accuracies and the final span-level exact match score values.

\subsubsection{Passage Length Distribution}
\begin{figure*}[h!]
\centering
\includegraphics[width=\linewidth]{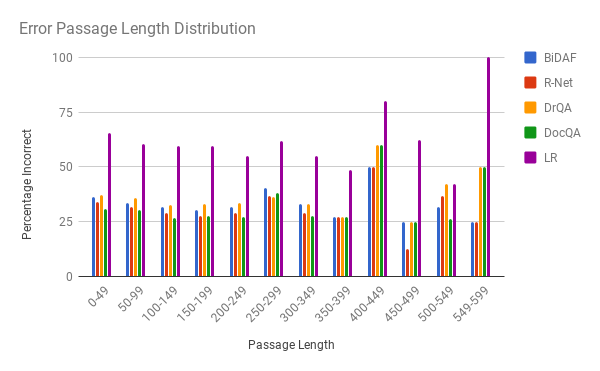}
\caption{Percentage of total QA pairs for each range of passage lengths which have incorrect predictions by different models}
\label{dist-err-pass}
\end{figure*}

We analyze the impact of passage length on errors, since this can be an important factor in determining the difficulty of understanding the passage. As seen in Figure \ref{dist-err-pass}, DocQA performs the best on shorter passages, while R-Net and BiDAF are observed to be better for longer passages. However, there are no systematic error patterns and overall error rates, surprisingly, are not much higher for longer passages. This means that predictions on long passages are almost as good as on short (presumably easier to understand) passages.

\subsubsection{Question Length Distribution}
\begin{figure*}[h!]
\centering
\includegraphics[width=\linewidth]{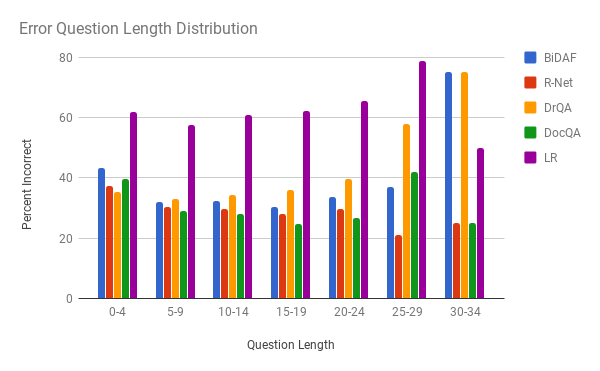}
\caption{Percentage of total QA pairs for each range of question lengths which have incorrect predictions by different models}
\label{dist-err-ques}
\end{figure*}

We also do a similar error analysis for questions of different lengths. Since there are very few questions which have length greater than 30, the estimate for range 30-34 is not very reliable. In Figure \ref{dist-err-ques}, we observe that the error rate first decreases and then increases for BiDAF, DrQA and DocQA. A plausible explanation for this is that shorter questions contain insufficient information in order to be able to select the correct answer span and can hence be confusing, but it also becomes difficult for end-to-end neural models to learn a good representation when the question becomes longer and syntactically more complicated. However, R-Net has an irregular trend with respect to question length, which is difficult to explain.

\subsubsection{Answer Length Distribution}
\begin{figure*}[h!]
\centering
\includegraphics[width=\linewidth]{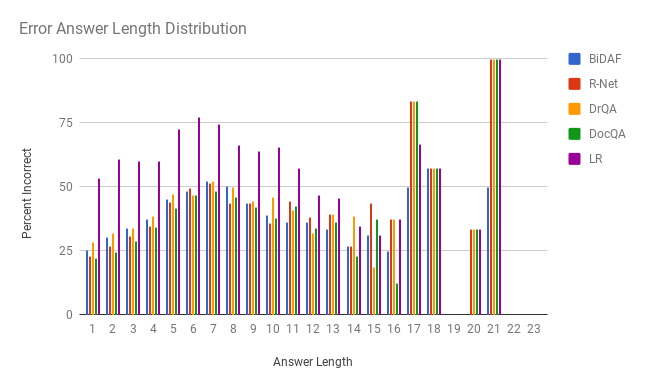}
\caption{Percentage of total QA pairs for each answer length which have incorrect predictions by different models}
\label{dist-err-ans}
\end{figure*}

For answers of varying lengths, the error rates are shown in Figure \ref{dist-err-ans}. Again, estimates for answers with length $>$16 are not very reliable since data is sparse for high answer lengths. Here, we observe an increasing trend initially and then a slight decrease (bell shape). This conforms to the hypothesis that shorter answers are easier to predict than longer answers, but only up to a certain answer length (observed to be around 7 for most models). The slightly better performance for very long answers is likely due to such answers having a higher chance of being (almost) entire sentences with simpler questions being asked about them.

\subsubsection{Error Overlap}

% \paragraph{Correct Answers}
% In Table \ref{eval-expt-corr}
% \begin{table}[ht!]
% \centering
% \resizebox{\linewidth}{!}{
% \begin{tabular}{|c|c|c|c|c|c|}
% \hline
% \bf{Model} & \bf{BiDAF} & \bf{R-Net} & \bf{DrQA} & \bf{DocQA} & \bf{LR}  \\ \hline
% \bf{BiDAF} &  7153  &  6317  &  5944  &  6394  &  3636 \\
% \bf{R-Net}  &  6317  &  7412  &  6150  &  6667 & 3726 \\
% \bf{DrQA} &  5944  &  6150  &  6976  &  6188 & 3555 \\
% \bf{DocQA} & 6394 & 6667 & 6188 & 7568 & 3734 \\
% \bf{LR} & 3636 & 3726 & 3555 & 3734 & 4243 \\
% \hline
% \end{tabular}
% }
% \caption{Correct Answer Overlap}
% \label{eval-expt-corr}
% \end{table}

% \paragraph{Incorrect Answers}
In Table \ref{eval-expt-incorr}, we analyze the number of erroneous predictions which overlap for different pairs of models, i.e., which belong to the intersection of the sets of incorrect answers generated by models in each (row, column) pair. Thus, the values in the table represent a symmetric matrix with diagonal elements indicating the number of errors which each model commits. This analysis can be useful while determining suitable models for creating meta ensembles since a low incorrect answer overlap indicates that the combined predictive power of the pair of models under consideration is high. We observe that most overlap values are in the range 20-25\% indicating that an ensemble might give considerably better performance than individual models. DocQA paired with other models generates low values, as expected, but the least value is observed for the DocQA-DrQA pair probably because they both use very different feature representations and architectures, and hence generate diverse outputs. Note that DrQA is not the second best performing model (among the ones we analyzed) when considered independently, but might add more value to an ensemble because of the observed answer overlap trends.

\begin{table}[ht!]
\centering
\resizebox{\linewidth}{!}{
\begin{tabular}{|c|ccccc|}
\hline
\bf{Model} & \bf{BiDAF} & \bf{R-Net} & \bf{DrQA} & \bf{DocQA} & \bf{LR}  \\ \hline
\bf{BiDAF} &  32.33  &  21.97  &  22.56  &  21.22  &  26.58 \\
\bf{R-Net}  &  21.97  &  29.88  &  22.06  &  21.35 & 24.99 \\
\bf{DrQA} &  22.56  &  22.06  &  34.00  &  \textbf{20.95} & 27.49 \\
\bf{DocQA} & 21.22 & 21.35 & \textbf{20.95} & 28.40 & 23.59 \\
\bf{LR} & 26.58 & 24.99 & 27.49 & 23.59 & 59.86 \\
\hline
\end{tabular}
}
\caption{Incorrect Answer Overlap (\%)}
\label{eval-expt-incorr}
\end{table}

\noindent One way in which this analysis can help in exploring ensemble-based methods is that instead of trying all possible combinations of models, we can adopt a greedy approach based on the incorrect answer overlap metric to decide which model to add to the ensemble (and only if it leads to a statistically significant difference in this overlap). After determining an approximately optimal set of models which such an ensemble should be composed of, each of these models can be trained independently followed by multi-label classification (to select one of the generated answers) using techniques like logistic regression, a feed-forward neural network or a recurrent or convolutional neural network with input features based on the question, the passage and their token overlap. The entire network can also be trained end-to-end.

%Also, all 5 models combined have an error overlap of 13.68\%, which means that if we had a mechanism to perfectly choose between these models depending on the question and passage, we would get an Exact Match score of 86.32\%, so this direction might be worth investigating. 

Also, all 5 models combined have an error overlap of 13.68\%, i.e., if we had a mechanism to perfectly choose between these models, we would get an Exact Match score of 86.32\%. This indicates that future work based on ensembling different neural models can give promising results and is worth exploring.
%which means that if we had a mechanism to perfectly choose between these models depending on the question and passage, we would get an Exact Match score of 86.32\%, so this direction might be worth investigating. 

An example of a passage-question-answer that all of the models get wrong is:

\noindent \textbf{Passage:} The University of Warsaw was established in 1816, when the partitions of Poland separated Warsaw from the oldest and most influential Polish academic center, in Krakow. Warsaw University of Technology is the second academic school of technology in the country, and one of the largest in East-Central Europe, employing 2,000 professors. Other institutions for higher education include the Medical University of Warsaw, the largest medical school in Poland and one of the most prestigious, the National Defence University, highest military academic institution in Poland, the Fryderyk Chopin University of Music the oldest and largest music school in Poland, and one of the largest in Europe, the Warsaw School of Economics, the oldest and most renowned economic university in the country, and the Warsaw University of Life Sciences the largest agricultural university founded in 1818. \\
\textbf{Question:} What is one of the largest music schools in Europe? \\
\textbf{Answer:} Fryderyk Chopin University of Music \\

\noindent This passage-question-answer is difficult for automatic processing because there several entities of the same type (school / university) in the passage, and the question is a paraphrase of one segment of a very long, syntactically complicated sentence which contains the information required to be able to infer the correct answer. This presents an interesting challenge, and such qualitative observations can be used to formulate a general technique for effectively testing machine reading systems.

% \subsubsection{Question Types}

% \paragraph{Initial Unigram}

% \paragraph{Initial Bigram}

\subsection{Qualitative Analysis}
For qualitative error analysis, we sample 100 incorrect predictions (based on EM) from each model and try to find common error categories. Broadly, the errors observed were either because of incorrect answer span boundaries or inability to infer the meaning of the question / passage. Examples of each error type are shown in Table \ref{examples-test}, and these are further described below.
\begin{table*}
\small
\resizebox{\textwidth}{!}{
\begin{tabular}{|p{0.1\linewidth}|p{0.48\linewidth}|p{0.24\linewidth}|p{0.18\linewidth}|}
\hline
\textbf{Error Type} & \textbf{Passage} & \textbf{Question} & \textbf{Predicted Answer}  \\ 
\hline
Incorrect answer boundary (longer) & ... survey of 4,745 North American Lutherans aged 15-65 found that, compared to the other minority groups under consideration, Lutherans were the \textcolor{blue}{least prejudiced} toward Jews. Nevertheless, Professor Richard (Dick) Geary, ... & What did a survey of North American Lutherans find that Lutherans felt about Jews compared to other minority groups? & 15-65 found that, compared to the other minority groups under consideration, Lutherans were the least prejudiced toward Jews \\ \hline
Incorrect answer boundary (shorter) & ... In the United States, in order for a prescription for a controlled substance to be valid, \textcolor{blue}{it must be issued for a legitimate medical purpose by a licensed practitioner acting in the course of legitimate doctor-patient relationship}. The filling ... & What conditions must be met to prescribe a controlled substance? & issued for a legitimate medical purpose \\ \hline
Soft Correct & ... for that time. The vBNS installed \textcolor{blue}{one of the first ever production OC-48c (2.5 Gbit/s) IP links} in February 1999 and went on to upgrade the entire backbone ... & What did the network install in 1999? & OC-48c (2.5 Gbit/s) IP links \\ \hline
Multi-Sentence & ... \textcolor{blue}{User Datagram Protocol} (UDP) is an example of a datagram protocol. In the virtual call system ... model. The X.25 protocol suite uses this network type. & X.25 uses what type network type? & protocol suite \\ \hline
Paraphrase & ... rather than consumers. There is \textcolor{blue}{no known case} of any U.S. citizens buying Canadian drugs for personal use with a prescription, who has ever been charged by authorities. & Has there ever been anyone charged with importing drugs from Canada for personal medicinal use? & has ever been charged by authorities \\ \hline
Same Entity Type / Unit Confusion & ... after the 1973 oil crisis, Honda, Toyota and Nissan, affected by the 1981 voluntary export restraints, opened US assembly plants and established their luxury divisions (Acura, \textcolor{blue}{Lexus} and Infiniti, respectively) to distinguish themselves from their mass-market brands. & Name a luxury division of Toyota. & Acura, Lexus and Infiniti \\ \hline
Requires World Knowledge & ... disobedience in opposition to the decisions of non-governmental agencies such as trade unions, banks, and \textcolor{blue}{private universities} can be justified if ... & What public entity of learning is often target of civil disobedience? & governmental \\ \hline
Missing Inference & ... \textcolor{blue}{Killer T cells} are a sub-group of T cells that kill cells that are infected with viruses (and other pathogens), or are otherwise damaged or dysfunctional. As with B cells ... & What kind of T cells kill cells that are infected with pathogens? & sub-group \\ \hline
\end{tabular}
}
\caption{Examples of error types observed in the qualitative analysis - \textcolor{blue}{blue} indicates ground truth}
\label{examples-test}
\end{table*}
\subsubsection{Boundary-Based Errors}
\paragraph{Incorrect answer boundary (longer):}
This error category includes those cases where the predicted span is longer than the ground truth answer, but contains the answer.
\paragraph{Incorrect answer boundary (shorter):}
This error category includes those cases where the predicted span is shorter than the ground truth answer, and is a substring of the answer.
\paragraph{Soft Correct:}
This error category includes those cases where the prediction is actually correct, but due to inclusion / exclusion of certain question terms (such as units) along with the answer, it is deemed incorrect.

\subsubsection{Inference-Based Errors}
\paragraph{Multi-Sentence:}
This error category includes those cases where inference is required to be performed across 2 or more sentences in the given passage to be able to arrive at the answer, which leads to an incorrect prediction based on only 1 passage sentence.
\paragraph{Paraphrase:}
This error category includes those cases where the question paraphrases certain parts of the sentence that it is asking about which makes lexical pattern matching difficult and leads to errors in prediction.
\paragraph{Same Entity Type Confusion / Unit Confusion:}
This error category includes those cases where the question is about an entity type which is present multiple times in the passage and the model returns a different entity than the ground truth entity but of the same type.
\paragraph{Requires World Knowledge:}
This error category includes questions which can not be answered using the given passage alone and require external knowledge to solve, leading to incorrect predictions.
\paragraph{Missing Inference:}
This category includes inference-related errors which don't belong to any of the other categories mentioned above.

\subsubsection{Observations}
\begin{figure*}[h!]
\centering
\includegraphics[width=\linewidth]{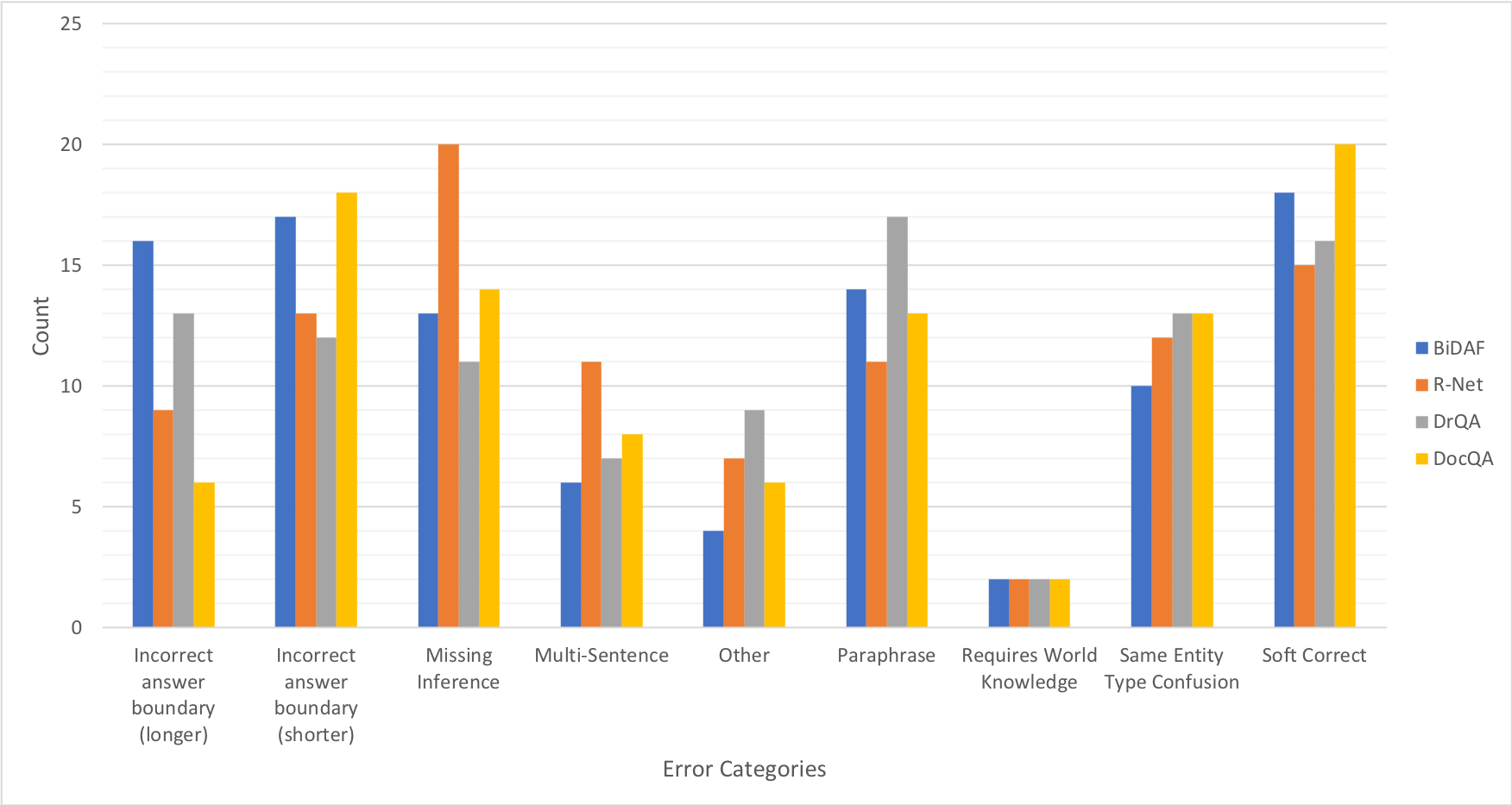}
\caption{Distribution of errors by various models across different categories using manual inspection}
\label{dist-err}
\end{figure*}

%\paragraph{BiDAF:}
In this section, we record the main observations from our qualitative error analysis and analyze potential reasons for the error trends observed. Figure \ref{dist-err} shows the different types of errors in predictions by various models.

We observe that BiDAF makes many boundary-based errors which indicates that a better output layer (since this is responsible for span identification -- although errors might have percolated from previous layers, most of these are cases where the model almost got the correct answer but not exactly) or some post-processing of the answer might help improve performance. Paraphrases also contribute to almost 15\% of errors observed which indicates that the question and the relevant parts of the context are not effectively matched in these cases.

% \begin{figure}[h!]
% \centering
% \includegraphics[width=\linewidth]{bidaf}
% \caption{Distribution of errors by BiDAF across different categories using manual inspection}
% \label{dist-err-bidaf}
% \end{figure}

%\paragraph{R-Net:}
%Figure \ref{dist-err-all} shows the different types of errors in R-Net predictions. 
We observe that R-Net makes fewer boundary errors, perhaps because self-attention enables it to accumulate evidence and return better answer spans, although this leads to more errors of the `shorter' answer type than `longer'. Also, missing inference contributes to almost 20\% of the observed errors (not including multiple sentences or paraphrases).

% \begin{figure}[h!]
% \centering
% \includegraphics[width=\linewidth]{rnet}
% \caption{Distribution of errors by R-Net across different categories using manual inspection}
% \label{dist-err-rnet}
% \end{figure}

%\paragraph{DrQA:}
%Figure \ref{dist-err-all} shows the different types of errors in DrQA predictions. 
Paraphrasing is the most frequent error category observed for DrQA, which makes sense if we consider the features used to represent each passage, such as exact match with a question word, which depend on lexical overlap between the question and passage.

% \begin{figure}[h!]
% \centering
% \includegraphics[width=\linewidth]{drqa}
% \caption{Distribution of errors by DrQA across different categories using manual inspection}
% \label{dist-err-drqa}
% \end{figure}

%\paragraph{DocQA:}
%Figure \ref{dist-err-all} shows the different types of errors in DocQA predictions. 
We observe that DocQA makes many boundary errors too, again making more mistakes by predicting shorter answers than expected in most of the observed cases. A better root cause analysis can be performed by visualizing outputs from different layers and evaluating these, and we leave this in-depth investigation to future work. Also, the high number of Soft Correct outputs across all models points to some deficiencies in the SQuAD annotations, which might limit the reliability of the performance evaluation metrics. % Thus, more sophisticated prediction layer(s) should help improve the performance even further.

Although these state-of-the-art deep learning models for machine reading are supposed to have inference capabilities, our error analysis above points to their limitations. These insights can be useful for developing benchmarks and datasets which enable realistic evaluation of systems which aim to `solve' the RC task. In \citet{2018arXiv180503830W}, we take a first step in this direction by proposing a method focused on questions involving referential inference, a setting to which these models fail to generalize well.

% \begin{figure}[h!]
% \centering
% \includegraphics[width=\linewidth]{docqa}
% \caption{Distribution of errors by DocQA across different categories using manual inspection}
% \label{dist-err-docqa}
% \end{figure}

\section{Conclusion and Future Work}
\label{sec:concl} 
In this work, we analyze - both quantitatively and qualitatively - results generated by 4 end-to-end neural models on the Stanford Question Answering Dataset. We observe interesting trends in the analysis, with some error patterns which are consistent across different models and some others which are specific to each model due to their different input features and architectures. This is important to be able to interpret and gain an intuition for the effective functions that different components in a neural model architecture perform versus their intended functions, and also to understand model-specific biases. Eventually, this can enable us to come up with new models including specific components which tackle these errors. Alternatively, the overlap analysis demonstrates that learning ensembles of different neural models to combine their individual strengths and quirks might be an interesting direction to explore to achieve better performance.

Even though the scope of this paper is restricted to SQuAD, similar analysis can be done for any datasets / models / features, to gain a better understanding and enable a better assessment of state-of-the-art in neural machine reading. To this end, we also performed some preliminary experiments on TriviaQA so as to analyze the difference between the properties of the two datasets, but were unable to replicate the published results owing to pre-processing / hyperparameters. We will continue to work on this since the ability of a model to generalize and to be able to learn from a particular domain and transfer some knowledge to a different domain is a very exciting research area.

We also believe that such analysis can help curate datasets which are better indicators of the actual natural language `reading' and `comprehending' capabilities of models rather than falling prey to shallow pattern matching. One way to achieve this is by building new challenges that are specifically designed to put pressure on the identified weaknesses of neural models. Thus, we can move towards the development of datasets and models which truly push the envelope of the challenging machine reading task.

\section*{Acknowledgments}
We would like to thank Chaitanya Malaviya and Abhishek Chinni for their valuable feedback, and the Language Technologies Institute at CMU for the GPU resources used in this work. We are also very grateful to the anonymous reviewers for their insightful comments and suggestions, which helped us polish the presentation of our work.

% include your own bib file like this:
%\bibliographystyle{acl}
%\bibliography{acl2018}
\bibliography{acl2018}
\bibliographystyle{acl_natbib}

\end{document}